%% file: domainadapt_iccvw23.tex
\documentclass[10pt,twocolumn,letterpaper]{article}

\usepackage{iccv}
\usepackage{times}
\usepackage{epsfig}
\usepackage{graphicx}
\usepackage{amsmath}
\usepackage{amssymb}

\usepackage{comment}

\iccvfinalcopy % *** Uncomment this line for the final submission

\def\iccvPaperID{0024} % *** Enter the ICCV Paper ID here

\ificcvfinal\fi

\usepackage{graphicx}
\usepackage{amsmath}
\usepackage{amssymb}
\usepackage{booktabs}

\usepackage{tikz}
\usetikzlibrary{shapes.geometric}

\usepackage{tabularx}
\usepackage{multirow}

\usepackage[utf8]{inputenc} % allow utf-8 input
\usepackage[T1]{fontenc}    % use 8-bit T1 fonts
\usepackage{url}            % simple URL typesetting
\usepackage{booktabs}       % professional-quality tables
\usepackage{amsfonts}       % blackboard math symbols
\usepackage{nicefrac}       % compact symbols for 1/2, etc.
\usepackage{microtype}      % microtypography
\usepackage{xcolor}         % colors
\usepackage{xspace}
\usepackage{graphicx}
\usepackage{tabularx}
\usepackage{tikz}
\usepackage{pifont} 

\usepackage{xfrac}
\usepackage{commath}
\usepackage{enumitem}
\setlist{leftmargin=5.5mm}
\usepackage{array}
\usepackage{algorithm}
\usepackage{algpseudocode}

\usepackage{calrsfs}
\DeclareMathAlphabet{\pazocal}{OMS}{zplm}{m}{n}
\newcommand{\Lb}[1]{\pazocal{#1}}

\newcolumntype{Y}{>{\raggedleft\arraybackslash}X}

\usepackage{comment}
\usepackage{amsmath,amssymb}
\usepackage{subcaption}
\usepackage{sidecap}
\usepackage{cite}

\input{definitions}

\usepackage[pagebackref=true,breaklinks=true,letterpaper=true,colorlinks,bookmarks=false]{hyperref}

\usepackage[capitalize]{cleveref}
\crefname{section}{Sec.}{Secs.}
\Crefname{section}{Section}{Sections}
\Crefname{table}{Table}{Tables}
\crefname{table}{Tab.}{Tabs.}

\begin{document}

%%%%%%%%% TITLE
%\title{Leveraging Cross-Modal Cues For Mitigating Data Paucity Challenges in Object Detection}
\title{Tensor Factorization for Leveraging Cross-Modal Knowledge in Data-Constrained Infrared Object Detection}
\author{Manish Sharma$^{1*}$ \qquad Moitreya Chatterjee$^{2}\thanks{Equal Contributions. Work done while interning at MERL.}$ \qquad Kuan-Chuan Peng$^{2}$  \qquad Suhas Lohit$^{2}$ \qquad Michael Jones$^{2}$\\
$^1$Rochester Institute of Technology, NY 14623, USA\\
$^2$Mitsubishi Electric Research Laboratories,
Cambridge, MA 02139, USA\\
\small{\texttt{ms8515@rit.edu\quad metro.smiles@gmail.com\quad kpeng@merl.com\quad slohit@merl.com\quad mjones@merl.com}}
}

\maketitle

%\begin{abstract}

%\end{abstract}

\input{abstract}
\input{intro}
\input{related_work}

\input{method}
\input{expts}

\input{conclude}

%%%%%%%%% REFERENCES
{\small
\bibliographystyle{ieee_fullname}
\bibliography{references}
}

\end{document}

%% file: definitions.tex
\newcommand{\name}{TensorFact\xspace}

%% file: abstract.tex
\begin{abstract}
%\vspace*{-0.5cm}

%\SL{The title is not mentioning tensor factorization or infrared. Should we have something like Tensor Factorization for Leveraging Cross-Modal Knowledge in Data-Constrained Infrared Object Detection?} 

%\SL{There are no references throughout the paper} 

%\SL{Related work section is empty} 
\vspace{-0.15cm}
While state-of-the-art object detection methods have reached some level of maturity for regular RGB images, there is still some distance to be covered before these methods perform comparably on Infrared (IR) images. The primary bottleneck towards accomplishing this goal is the lack of sufficient labeled training data in the IR modality, owing to the cost of acquiring such data. Realizing that object detection methods for the RGB modality are quite robust (at least for some commonplace classes, like person, car, etc.), thanks to the giant training sets that exist, in this work we seek to leverage cues from the RGB modality to scale object detectors to the IR modality, while preserving model performance in the RGB modality. At the core of our method, is a novel tensor decomposition method called \emph{\name} which splits the convolution kernels of a layer of a Convolutional Neural Network (CNN) into low-rank factor matrices, with fewer parameters than the original CNN. We first pre-train these factor matrices on the RGB modality, for which plenty of training data are assumed to exist and then augment only a few trainable parameters for training on the IR modality -- to avoid over-fitting, while encouraging them to capture complementary cues from those trained only on the RGB modality. We validate our approach empirically by first assessing how well our \emph{\name} decomposed network performs at the task of detecting objects in RGB images vis-\'a-vis the original network and then look at how well it adapts to IR images of the FLIR ADAS v1 dataset. For the latter, we train models under scenarios that pose challenges stemming from data paucity. From the experiments, we observe that: (i) \emph{\name} shows performance gains on RGB images; (ii) further, this pre-trained model, when fine-tuned, outperforms a standard state-of-the-art object detector on the FLIR ADAS v1 dataset by about $4\%$ in terms of mAP 50 score. 
%\MJ{The abstract is a little unclear about our contribution.  The beginning says that our goal is to improve object detection in the IR modality but then the end claims improvement in RGB.  Also, there should be a more direct explanation of why decomposition into several low-rank components is expected to improve IR detection.} %, underscoring the effectiveness of our method. 
%Moreover, our decomposition also shows performance gains on the RGB images itself, by helping to avoid overfitting.

%  from the FLIR-Aligned dataset

% Furthermore, the shift from RGB modality to IR, manifests in such a sharp change in the input space that domain adaptation methods are unable to handle such a shift. 

% Most prior methods design complex algorithmic techniques to mitigate this challenge but achieve limited success in doing so.

\end{abstract}

%% file: intro.tex
\section{Introduction}
\begin{figure}[t]
    \centering
  \includegraphics[width=1\linewidth, trim={9.2cm 3cm 0.3cm 3.4cm},clip]{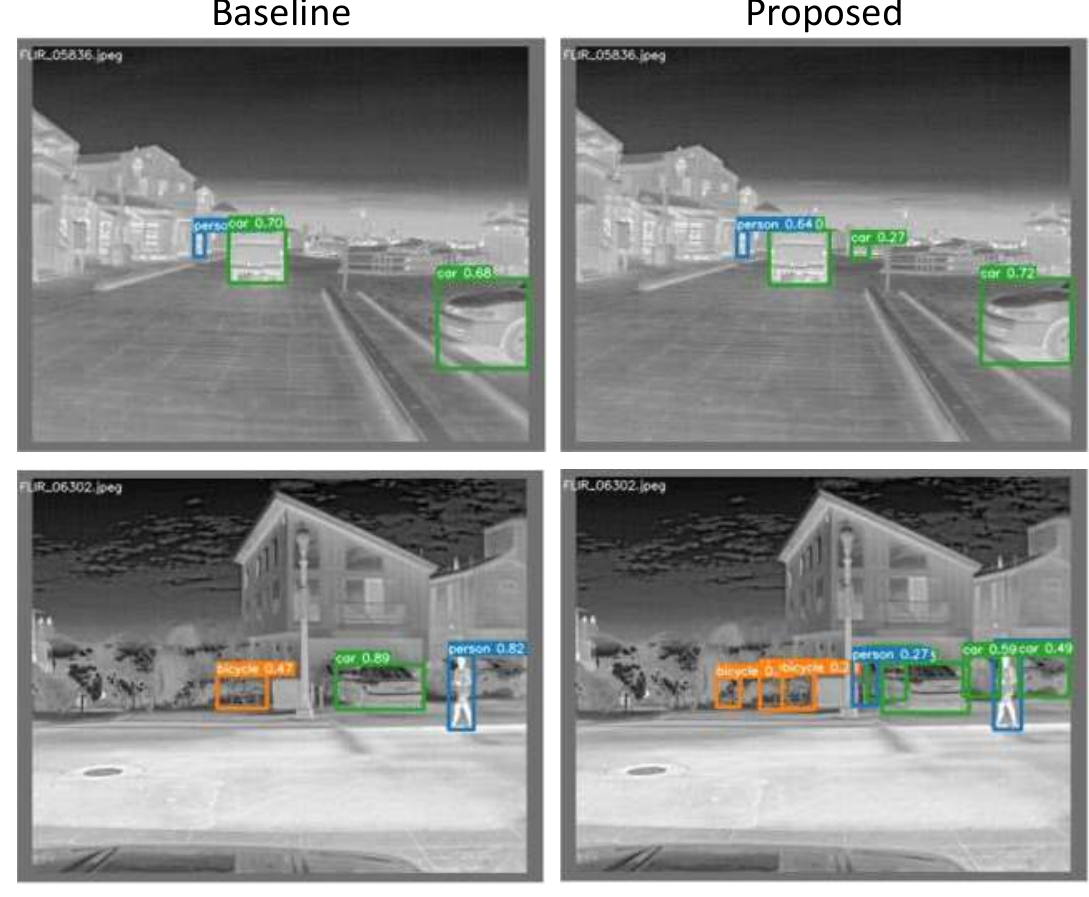}
  \caption{Qualitative comparison of object detections by a state-of-the-art object detector (denoted as baseline)~\cite{wang2023yolov7} and our \name method on IR images. The \textcolor{orange}{orange}, \textcolor{cyan}{cyan}, \textcolor{green}{green} boxes denote bicycle, person, and car classes respectively, while the associated numbers denote the confidence score of the prediction. The visualizations show that our proposed approach is better at capturing more objects, especially those that are of a smaller size, with higher precision.  } % \KP{Why there's no ground truth in this figure? If you'd like to show the colors of the bounding boxes, please explain the colors. (the same for Fig. 5 and 6.)}
  % \SL{Image is wrong}
  \label{fig:task}
%\vspace{-6mm}
\end{figure}

The success of deep neural networks in core computer vision tasks, such as image denoising \cite{sharma2021convolutional}, image classification \cite{singh2023multimodal}, object detection \cite{dhanaraj2020vehicle, sharma2020yolors}, \etc. can at least in part be attributed to the availability of large-scale labeled training data~\cite{sun2017revisiting}, which allows these models (with lots of parameters) to avoid over-fitting~\cite{dubey2018coreset}. This has resulted in wide-ranging applicability of these methods in tasks such as pedestrian detection in vehicles~\cite{ouyang2013joint}, face detection~\cite{sun2015deepid3}, vehicle counting~\cite{zhang2017fcn}, \etc.

One key element that made such large-scale data available, is the ubiquity of good quality RGB cameras, which come at throwaway prices. 
%Thanks to the mass adoption of smartphones, the count of which currently exceeds yyy billion globally ~\cite{}, these RGB cameras abound, resulting in lots of RGB images \SL{Datasets in autonomous driving are not because of smartphones, better to say that you get very good quality images for low cost}. 
This coupled with the popularity of online platforms for sharing content widely, including social media sites such as YouTube or Meta, meant that sharing such images at a large-scale became commonplace. 

However, from the standpoint of certain applications, such as autonomous driving, regular RGB images fall short on some important counts. For instance, while RGB images can provide clear visualization of the surroundings during the day, at night, RGB images are only useful if there is sufficient street lighting, \etc. In scenarios where the ambient light is insufficient, passive thermal Infrared (IR) cameras come in handy for tasks such as pedestrian detection, as thermal IR sensors capture scenes at wavelengths beyond the visible spectrum and are sensitive to warm objects, such as the human body~\cite{campbell2002biological}. Nonetheless, one catch that remains is that IR cameras are not as cheap as their RGB counterparts and are thus not as ubiquitous. This poses a major hurdle in acquiring the profuse amounts of images needed to train deep networks that could operate on IR images at performance levels similar to their RGB counterparts. {In such conditions, an overparameterized model results in overfitting, which has an impact on model generalisation and performance. Therefore, a reduction in the number of parameters may be needed for improved performance. Low-rank factorization methods are among the most popular methods towards this end and are utilized for different deep learning applications \cite{sharma2021yolors, hu2021lora, kamalakara2022exploring}.}

While the success of deep neural networks today spans several computer vision tasks, the task of object detection is of particular interest in this paper.
%, and forms an initial step for undertaking further downstream tasks. 
The task entails localizing the pixels which an object occupies in an image as well as labeling the cluster of pixels with the class to which the said object belongs. Solving this task is crucial, since it permits acquiring a greater understanding of what an image contains and is often a first step towards understanding the scene~\cite{li2017scene}. Given the importance of IR images, as a modality for the task of scene understanding, designing effective object detection models that work on such data becomes critical. Nonetheless, the paucity of sufficient training data (\ie, datasets with lots of IR images) continues to present a challenge to this end.

In this work, we leverage the observation that while sufficient training data in the IR modality may be difficult to find, such data for the RGB modality is easily available. The key idea in our approach then, is to train an object detection model in the RGB modality and to then transfer the common cross-modal cues to the IR modality where only a few parameters can be trained to capture the complementary cues necessary for successfully detecting objects in the IR image space. Concretely, we devise a novel method called \emph{\name}, which splits the convolution kernel weights of a CNN layer into low-rank factor matrices, with fewer trainable parameters. These factor matrices can be trained to capture the common cues for detecting objects, across modalities, by leveraging the RGB data. These weights can then be augmented with only a few, new learnable parameters to capture the cues specific to the IR modality. This design allows us to train only the relatively small number of IR modality-specific weights when training with IR images, allowing us to prevent over-fitting. 
%To the best of our knowledge this cross-modal setup for object detection (RGB to IR), has not been looked at before and 
Note that na\"ively applying domain adaptation methods~\cite{ahmed2022cross} to transfer from RGB to IR modality fails because here the modality itself switches between the source (RGB) and the target (IR) which represents a big shift in the data distribution.
% \SL{Not sure if this line is necessary}.

We conduct experiments on the FLIR ADAS v1 dataset~\cite{flir_adas_v1} of IR images to empirically validate the efficacy of our method. To derive the common object detection cues from RGB images, we use the FLIR Aligned RGB~\cite{flir_aligned} images. Our experiments show that \emph{\name} decomposition assists with achieving better object detection performance both on RGB and IR images, even when the latter has few training samples. In particular, in the IR dataset (FLIR ADAS v1), our method outperforms a competing state-of-the-art object detection model~\cite{wang2023yolov7} by $4\%$ on mAP 50, underscoring the efficacy of our method. Figure~\ref{fig:task} contrasts detections obtained by our method in comparison to a recent state-of-the-art detection baseline, YOLOv7~\cite{wang2023yolov7}, on the FLIR ADAS v1 dataset. From the figure, we see that our approach is more capable of detecting objects of different sizes, compared to the state-of-the-art approach. 

We summarize below the core contributions of our work. %\KP{Don't we want to emphasize that \name outperforms the SOTA in the low-data regime in the contribution paragraph? After all, that's the topic of the workshop and this paper.}
\begin{itemize}
    \item  We present \emph{\name}, a novel tensor decomposition-based method that can leverage both modality-specific and cross-modal cues for effective object detection in the IR modality, where acquiring sufficient training data is a challenge. 

    \item Our experiments reveal that our proposed method outperforms competing approaches at the task of object detection in a data sparse IR modality, with only 62 training images, by $4\%$ on mAP 50.
    
    \item {Our formulation also offers a supplementary contribution to the RGB modality, yielding a compressed neural network that improves object detection in this modality.}
    %This area represents a promising direction for further exploration and comparison with major baselines.}

    % Our formulation yields a compressed neural network for detecting objects in the RGB modality, while also improving the object detection performance in the said modality. 
    %is generic enough where a user could plug in his/her desired priors to capture cross-modmodel architecture is simple while being effective, where the complementary, modality-specific features are learnt using easy to implement loss terms, such as $L_1$ loss. \SL{I don't think this is a contribution}.

\end{itemize}

%\begin{figure*}[t]
%\begin{center}
%  %\includegraphics[width = .9\linewidth]{figs/system_overview.png} 
%  %AVLEN-NeurIPS/images_neurips/architecture-cropped.pdf
%  %\includegraphics[width = .9\linewidth]{figs/archi.png} 
%  \includegraphics[width = 1\linewidth]{figs/arch_overview2.png}
%  % \includegraphics[width = 1\linewidth]{figs/Arch_Fig_v5.pdf} 
%  \caption{ Architecture of our \name pipeline. We show the bi-level  RL policies that the model learns, namely a top-level policy $\pi_s$ and three low-level policies $\pi_g,\pi_l$, and $\pi_{ques}$. We also show the various input modalities and the control flow.}
%  \label{fig:model1}
%  \end{center}
%\vspace{-7mm}
%\end{figure*}

%% file: related_work.tex
\section{Related works}
In this section, we discuss relevant prior works to our paper and present the distinction between these approaches and our method.

\noindent \textbf{Object detection approaches in IR images:}
{The journey of object detection in RGB images, using deep learning, has come a long way \cite{ren2015faster,redmon2016you, sharma2021yolors, wang2023yolov7}. 
The inception of a two-stage object detection process involving proposal generation and object class prediction, initiated by the work of Girshick \etal~\cite{girshick2014rich} for RGB images, laid the foundation for the field. However, the computational intensity of the process necessitated faster successors \cite{girshick2015fast, ren2015faster, ustinova2016learning, he2017mask, sun2021sparse}.
% , which came in the form of Fast R-CNN \cite{girshick2015fast} and Faster R-CNN \cite{ren2015faster}, utilizing Region of Interest (ROI) pooling and a novel Region Proposal Network (RPN).
%, particularly in the context of infrared (IR) images,
% Several solutions built upon the Faster R-CNN framework have emerged to tackle specific challenges, including R-FCN \cite{ustinova2016learning}, Mask R-CNN \cite{he2017mask}, and Sparse R-CNN \cite{sun2021sparse}. 
However, porting these approaches to the realm of IR image object detection, has posed certain challenges. The study by Ghose \etal~\cite{ghose2019pedestrian} and Devagupta \etal~\cite{devaguptapu2019borrow} sought to enhance infrared image features using saliency maps and multimodal Faster R-CNN, respectively. These efforts, however, encountered challenges such as slow inference speed, non end-to-end multitask training, and a lack of general applicability across different datasets.}
% , while Park \etal.~\cite{park2019cnn} focused on human detection in infrared images

To overcome the limitations of two-stage detectors, the work by Redmon and Farhadi~\cite{redmon2016you} introduced a one-stage detector, YOLO, which considered each image cell as a proposal for object detection and achieved end-to-end real-time detection. YOLO's evolution into YOLOv3~\cite{redmon2018yolov3}, YOLOv4~\cite{bochkovskiy2020yolov4}, and its subsequent variants, as documented by Kristo \etal~\cite{krivsto2020thermal}, has accelerated the detection of objects both in RGB and IR images, though issues of  omission of small-scale objects and low detection accuracy persist.

Innovative modifications like the SE block in SE-YOLO~\cite{li2020research} and the attention module, CIoU loss, improved Soft-NMS, and depthwise separable convolution in YOLO-ACN~\cite{li2020yolo} were proposed to improve detection accuracy, but they still grapple with challenges like large parameter sizes and applicability to embedded settings.

Other one-stage models have been explored, including ThermalDet~\cite{cao2019every} and TIRNet~\cite{dai2021tirnet}, each of which offers different solutions to the aforesaid problems but falls short when tested in real-world, non-curated datasets. Song \etal~\cite{song2021multispectral} proposed a multispectral feature fusion network based on YOLOv3, showing promise for smaller-sized images.

The YOLO series has shown considerable potential for IR object detection and several variants to it have been proposed. This includes the network of Shuigen \etal~\cite{shuigen2021infrared}, an attention mechanism-infused YOLOv3~\cite{ghose2019pedestrian}, and a YOLOv3 enhanced with a category balance loss term~\cite{li2021infrared}. Further refinements in object detection have been achieved by using the SAF architecture~\cite{manssor2022real} and the YOLO-FIRI model~\cite{li2021yolo}, which incorporate optimization parameters, introduce dilated convolutional block attention modules, and enable the detection of smaller IR targets. Zhao \etal~\cite{zhao2021iyolo} and Du \etal~\cite{du2021fa} have contributed to the field by improving the fusion method of YOLOv3 and leveraging YOLOv4 to enhance IR target features, respectively, paving a promising path for future IR object detection research. While we consider these  models for designing the backbone of our proposed approach but none of them provide a way to mitigate the data paucity issue in the IR modality which we address front and center.

\noindent \textbf{Domain adaptation methods:} The community has explored domain adaptation methods to overcome the challenges associated with less training data in certain domains. Towards this end, several works have been proposed~\cite{guan2021uncertainty,yao2021multi,rodriguez2019domain,zhang2022multi,wei2020incremental}, which include those that progressively transition from one domain to another~\cite{hsu2020progressive}, or transition through multiple levels of granularity~\cite{zhou2022multi}, or use semi-supervised~\cite{wang2023ssda3d,donahue2013semi} or unsupervised learning~\cite{yu2019unsupervised,li2021category} techniques for the same. Nonetheless, these approaches tackle scenarios which represent reasonably minor shifts in the domain of the input data, say from clear RGB images to foggy RGB images~\cite{erkent2020semantic} and so on. However, our task, deals with much larger-scale shifts in the type of input, in particular from RGB to IR modalities. The change is so stark that certain objects are visible in a given modality, only under specific scenarios. For instance, warm-bodied, dimly lit objects are visible only in the IR images but are very difficult to see in RGB images. This prevents us from trivially adapting these approaches for our task. While some more recent methods have looked into domain adaptation techniques for IR detection tasks, these are fairly limited in scope~\cite{kieu2020task,herrmann2018cnn} and focus mostly on detecting people, not other classes. Importantly, none of these approaches simulate the training data paucity scenario, for the IR modality, something we consider in this work.

%% file: method.tex
\section{Proposed approach}
\label{sec:method}

In this work, we propose \emph{\name} -- a novel tensor decomposition-based method designed to tackle the paucity of labeled training data in the IR modality. It effectively leverages knowledge learned from the RGB modality, where training data is abundant, and efficiently transfers this knowledge to the IR modality, overcoming the data scarcity challenge. Initially, we learn two trainable low-rank factor-matrices, the product of which yields the weights for each layer of the CNN and task them with detecting objects in the source RGB modality. This representation cuts down on the number of learnable parameters in the network and facilitates the training of a more generalizable network (due to less over-fitting) on the RGB modality. Following this, in order to facilitate object detection in the IR modality, we enhance the network's capability by a minor expansion of the number of trainable parameters. This is achieved by increasing the number of the columns/rows of the factor matrices. The factor matrices that emerge from the increased columns/rows effectively serve as a parallel trainable branch, enabling the network to leverage the complementary information gleaned from the RGB modality for object detection in the IR modality. In this way, \emph{\name} affords us a practical solution to the challenge of limited training data in the IR modality, demonstrating how robust and transferable features can be effectively extracted and utilized across different modalities.

\subsection{Notation}

In this paper, we utilize the following conventions: lowercase letters such as $x$ denotes scalar variables, vectors are symbolized by boldface lowercase letters like $\mathbf{x}$, and matrices are depicted by boldface uppercase letters such as $\mathbf{X}$. Tensors, on the other hand, are indicated by calligraphic uppercase letters (for instance, $\Lb{X}$). $\mathbb{R}$ denotes the set of real numbers. To illustrate a component of a vector, matrix, or tensor, we adopt the $[\cdot]_i$ notation, where $i$ represents a set of indices for that component.

\begin{figure}[t!]
    \centering
    \includegraphics[width=1\linewidth, trim={8.0cm 8.6cm 8.0cm 8.7cm},clip]{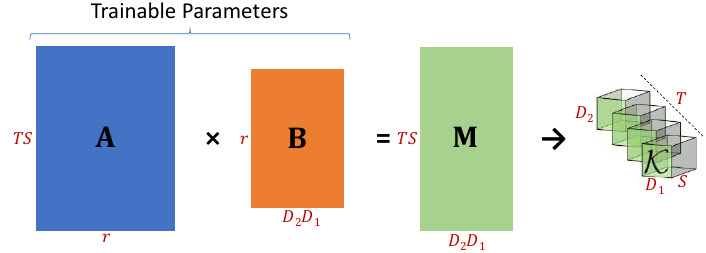}
    \caption{Decomposed convolutional layer. 
    %\KP{Please enlarge the text in Fig.2/3. Otherwise, the subscripts (\eg, $D_2$) are hard to read.}
    }
    \label{rgb_training}
\end{figure}

\subsection{Decomposed convolution layer}
The weights of a convolutional layer in a CNN, denoted by $\Lb{K} \in \mathbb{R}^{T \times S \times D_2 \times D_1}$, is a $4$-way tensor, where $D_1$ and $D_2$ represent the width and height, respectively, of the spatial window of the convolution kernels, while $S$ and $T$ denote the number of input channels of the input to the layer and the number of kernels learned in the layer. The number of trainable parameters in a standard convolutional layer is then given by $P = TSD_2D_1$.

For a decomposed convolutional layer, we commence with two trainable factors $\mathbf{A} \in \mathbb{R}^{TS \times r}$ and $\mathbf{B} \in \mathbb{R}^{r \times D_2D_1}$ with $r$ serving as their inner dimension, as shown in Figure \ref{rgb_training}, denoting the rank of the original weight matrix (prior to decomposition). These combine to form the intermediate matrix $\mathbf{M} = \mathbf{A}\mathbf{B}$, as follows:
\begin{equation}
[\mathbf{M}]_{p,q} = \sum_{c=1}^r [\mathbf{A}]_{p,c} [\mathbf{B}]_{c,q},
\label{eq:mat_decom}
\end{equation}
where, $p=1, \ldots, TS$ and $q=1, \ldots, D_2D_1$. This matrix $\mathbf{M}$, operates on the input to the layer. The convolutional filter $\Lb{K}$, is derived from $\mathbf{M}$ as:
\begin{equation}
[\Lb{K}]_{t,s,d_2,d_1} = [\mathbf{M}]_{(t-1)S+s,(d_2-1)D_1+d_1},
\label{eq:ten_mat}
\end{equation}
where, $t=1, \ldots, T$, $s=1, \ldots, S$, $d_2=1, \ldots, D_2$, and $d_1=1, \ldots, D_1$.
Therefore, the number of trainable parameters in the decomposed convolutional layer formulation, $P_{fac}$, is a function of $r$, resulting in $P_{fac} = r(TS + D_2D_1)$ trainable parameters. The value of $r$ can be altered to adapt to the necessary CNN complexity but typically $r \leq \text{rank}(\mathbf{M})$. Since CNNs are known to be over-parameterized~\cite{dubey2018coreset}, one could choose $r$ such that the number of learnable parameters is fewer than that in $\mathbf{M}$, to avoid the risk of over-fitting.

\subsection{Capacity augmentation}
\begin{figure}[t!]
    \centering
    \includegraphics[width=1\linewidth, trim={7.7cm 7.1cm 7.7cm 7.0cm},clip]{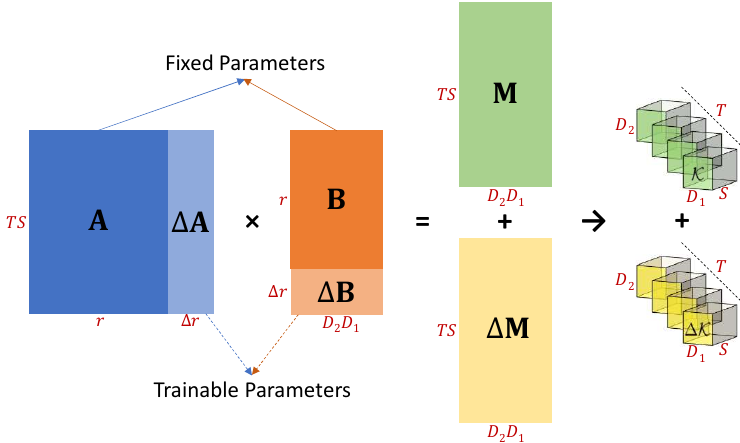}
    \caption{Decomposed convolutional layer with capacity augmentation. 
    %\KP{For the arrows in Fig. 3/4, could you please make them larger and more obvious (especially in Fig. 4 since the direction is important)?}
    }
    \label{ir_training}
\end{figure}

\begin{figure}[t!]
    \centering
    \includegraphics[width=0.7\linewidth, trim={7.4cm 9.5cm 9.0cm 8.7cm},clip]{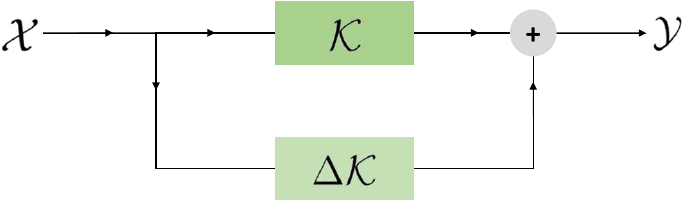}
    \caption{The flow of data in our proposed ~\emph{\name} approach in every layer. The input $\Lb{X}$ convolves with $\Lb{K}$ (top branch) and $\Delta{\Lb{K}}$ (bottom branch) and results in output $\Lb{Y}$ after summation.}
    \label{branches}
\end{figure}

To augment the network capacity to accommodate the new modality, we increase $r$ by $\Delta{r}$ (where $\Delta{r} > 0$) for both matrices $\mathbf{A}$ and $\mathbf{B}$, thereby producing $\mathbf{A'} \in \mathbb{R}^{TS \times (r+\Delta{r})}$ and $\mathbf{B'} \in \mathbb{R}^{(r+\Delta{r}) \times D_2D_1}$ with $r+\Delta{r}$ serving as their new inner dimension. Now, $\mathbf{A'}$ and $\mathbf{B'}$ can be interpreted as $\mathbf{A'} = \begin{bmatrix} \mathbf{A} \ || \Delta{\mathbf{A}} \end{bmatrix}$ and $\mathbf{B'} = \begin{bmatrix} \mathbf{B} \ || \Delta{\mathbf{B}} \end{bmatrix}^T$ such that $\Delta{\mathbf{A}} \in \mathbb{R}^{TS \times \Delta{r}}$ and $\Delta{\mathbf{B}} \in \mathbb{R}^{\Delta{r} \times D_2D_1}$ and $||$ denotes concatenation. Subsequently, $\mathbf{A'}$ and $\mathbf{B'}$ merge to form $\mathbf{M'} = \mathbf{A'}\mathbf{B'} = \mathbf{M} + \Delta{\mathbf{M}}$, where $\Delta{\mathbf{M}} = \Delta{\mathbf{A}}\Delta{\mathbf{B}}$, as shown in Figure \ref{ir_training}. Similar to Equation \ref{eq:ten_mat}, $\Delta{\Lb{K}} \in \mathbb{R}^{T \times S \times D_2 \times D_1}$ can be derived from $\Delta{\mathbf{M}}$. Hence, increasing $r$ by $\Delta{r}$ results in a parallel architectural branch, as depicted in Figure \ref{branches}. Therefore, the increase in the number of trainable parameters in a decomposed convolutional layer after capacity augmentation is given by $\Delta{P_{fac}} = \Delta{r}(TS + D_2D_1)$. We seek to augment as few parameters as possible to ensure the detection network does not suffer from challenges related to over-fitting in the new modality. In particular, we ensure that the total number of network parameters (considering those trained using only RGB and the augmented set) of our proposed framework, is less  than the original unfactorized network.
%\KP{What's the purpose of introducing the notations $P_{fac}$ and $\Delta{P_{fac}}$ when they are only mentioned once in the entire paper?}

\subsection{Training}

For an object detector CNN with $L$ convolutional layers, let $\mathbf{A}_l$ and $\mathbf{B}_l$ represent the left and right factor matrices, respectively, for the $l^{th}$ decomposed convolutional layer, with $r_l$ representing their inner-dimension and $l=1, \ldots, L$. When training for the data-rich source RGB modality, the network weights for the decomposed convolutional layers are SVD-initialized, leading to orthogonal column and row vectors in $\mathbf{A}_l$ and $\mathbf{B}_l$, respectively, with $r_l=\lfloor \alpha r_l^{max} \rfloor$. Here, $r_l^{max}=\min(TS,D_2D_1)_l$ and $\alpha \in (0, 1)$ controls the number of the trainable parameters across layers. With $\alpha \leq 1$, the training process is straightforward and similar to a typical object detector network, leading to the learning of both generic and modality-specific features for the RGB data.

Next, to train for the data-scarce IR modality, we augment the network capacity by increasing the value of $\alpha$, which introduces new trainable parameters and creates a parallel path for each decomposed convolutional layer. During this training phase, we freeze the trainable parameters learned during the training of the RGB modality, thereby architecturally promoting the learning of complementary features for the IR modality branch. Akin to skip-connections in ResNets~\cite{he2016deep}, which permits the learning of residual mapping, our proposed method leverages cross-modal cues and promotes the learning of features specific to the IR modality that were not learned during RGB modality training. As the factor matrices trained on the RGB data capture several cues essential for object detection, only a small percentage of augmented capacity is required for capturing the facets of object detection in the IR modality. This is an essential requirement to train the model without over-fitting in a data-scarce modality. Additionally, to explicitly capture complementary cues between the RGB and IR modalities, we maximize the $L_2$ or $L_1$ distances between the feature activation maps that are output from each branch (RGB and IR) of a layer and have it as an additional term in the training objective for the task. This can be implemented by the following loss $L_c$:
\begin{equation}
L_c = - || \Lb{K} * \Lb{X} - \Delta{\Lb{K}} * \Lb{X}||_p,
\label{eq:reg}
\end{equation}
where $p = \{1, 2\}$ and $*$ denotes convolution. Note that the dimensions of $\Lb{K}$ and $\Delta{\Lb{K}}$ are the same. The final loss function $L_f$ of \emph{\name} can be written as follows:
\begin{equation}
L_f = L_d + \omega_c L_c,
\label{eq:final}
\end{equation}
where $L_d$ is the object detection loss used in YOLOv7~\cite{wang2023yolov7}, and $\omega_c$ is the weight of $L_c$. We minimize this loss using the ADAM optimizer~\cite{kingma2014adam}.

%% file: expts.tex
\section{Experiments}
\label{sec:expts}

In this section we layout the empirical evaluation that we conducted to validate the efficacy of our proposed approach.

\begin{figure*}[t!]
    \centering
    \includegraphics[width=1\linewidth, trim={0.2cm 3cm 0.3cm 3.4cm},clip]{figs/results_1.pdf}
    \caption{Comparison of object detection results between the state-of-the-art YOLOv7~\cite{wang2023yolov7} and our proposed approach. We show the ground truth (left column), baseline (middle column), and proposed method's ($\alpha=0.1$, right column) detections as rectangular bounding boxes. We show detections on two different images from the FLIR ADAS v1 IR validation dataset, one in each row. The \textcolor{orange}{orange}, \textcolor{cyan}{cyan}, \textcolor{green}{green} boxes denote bicycle, person, and car classes respectively, while the associated numbers denote the confidence score of the prediction. 
    %\KP{Since we have more space available, could you please show more qualitative examples in both Fig. 5 and 6?}
    }
    \label{results_1}
\end{figure*}

\begin{figure}[t!]
    \centering
    \includegraphics[width=1\linewidth, trim={7.2cm 9.5cm 0.0cm 3.6cm},clip]{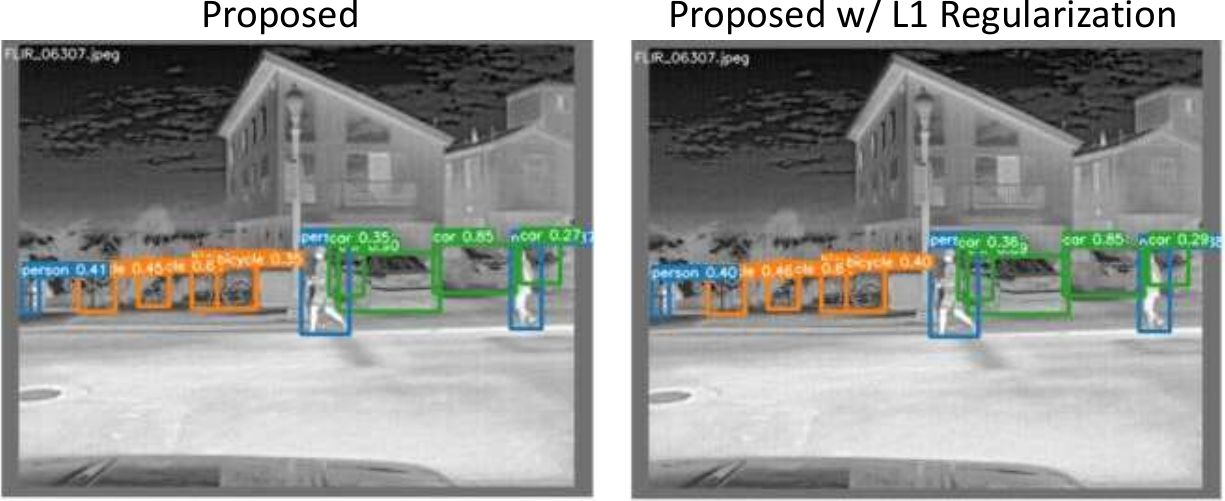}
    \caption{Comparison of object detection results for the proposed method without (left column) and with (right column) $L_1$ regularization. The \textcolor{orange}{orange}, \textcolor{cyan}{cyan}, \textcolor{green}{green} boxes denote bicycle, person, and car classes respectively, while the associated numbers denote the confidence score of the prediction. We obtain better object detections using the $L_1$ regularization, as compared to the vanilla model, as manifested by the higher confidence scores for the predicted bounding boxes.
    %\KP{Why there's no ground truth in this figure?}
    }
    \label{results_2}
\end{figure}

\subsection{Experimental setup}
\noindent \textbf{Datasets:} 
In our object detection experiments, we make use of two datasets: FLIR Aligned \cite{flir_aligned} and FLIR ADAS v1 \cite{flir_adas_v1}. The FLIR Aligned dataset contains RGB images, with ground-truth comprising bounding-box coordinates around objects in the image as well as class labels. This dataset includes $4129$ images for training and $1013$ images for validation, and features three classes: person, bicycle, and car, with the distribution of instances provided in Table \ref{tab:flir_aligned_dist}. 
%We use only the RGB modality from the FLIR Aligned dataset in our experiments.

The FLIR ADAS v1 dataset is a dataset of IR images.
%but it lacks spatial and temporal co-registration. 
The ground-truth for this dataset, 
%applicable only to the IR modality, 
includes bounding-box coordinates around objects in the image and their class labels, from among: person, bicycle, and car. The dataset includes $7859$ images for training and $1360$ images for validation. However, for fair comparative studies, we split the training set into train and validation splits in an 80:20 train-validation ratio to create new, randomly selected train and validation sets consisting of $6287$ and $1572$ images, respectively. To mimic a data-scarce environment, we use randomly selected $62$ images ($1\%$ of images) from the training set. Table \ref{tab:flir_adas_v1_dist} details the distribution of the FLIR ADAS v1 IR dataset classes, as used in our experiment.

\begin{table}[!t]
\centering
\footnotesize
%\fontsize{6}{8.5}\selectfont
%\begin{tabularx}{\linewidth}{XYY}
\begin{tabular}{ccc}
\toprule
\textbf{Class} & \textbf{Training Instances} & \textbf{Validation Instances}\\
\midrule
Person & $8987$ & $4107$\\
Bicycle & $2566$ & $360$\\
Car & $20608$ & $4124$\\
\bottomrule
\end{tabular}
\caption{Distribution of class instances for training and validation sets for FLIR Aligned RGB dataset~\cite{flir_aligned}.\label{tab:flir_aligned_dist}}
\end{table}

\begin{table}[!t]
\centering
\footnotesize
%\fontsize{6}{8.5}\selectfont
%\begin{tabularx}{\linewidth}{|X||Y||Y|}
\begin{tabular}{ccc}
\toprule
\textbf{Class} & \textbf{Training Instances} & \textbf{Validation Instances}\\
\midrule
Person & $161$ & $4611$\\
Bicycle & $24$ & $842$\\
Car & $351$ & $8472$\\
\bottomrule
\end{tabular}
\caption{Distribution of class instances for training ($1\%$) and validation sets for FLIR ADAS v1 IR dataset~\cite{flir_adas_v1}.\label{tab:flir_adas_v1_dist}}
\end{table}

\noindent \textbf{Baseline network and evaluation metrics:}
We use YOLOv7~\cite{wang2023yolov7}, a state-of-the-art object detector with over 37M trainable parameters as our baseline network. To determine appropriate anchor box sizes for the detector, we use K-Means++ method~\cite{arthur2007k}.

In evaluating the performance of our object detection model, we employed the Mean Average Precision (mAP), a widely-used and robust metric in the field. mAP considers both precision and recall, ensuring a balance between detecting as many objects as possible and minimizing false positives. This is achieved by generating Precision-Recall (PR) curves for each object class in two different settings. In the first, the Intersection Over Union (IoU) between predicted and ground truth bounding boxes is set to larger than 0.5, for it to be counted as a true positive prediction, while in the second setting, multiple evaluations are performed with increasing thresholds from 0.5 to 0.95 in increments of 0.05. The Average Precision (AP) is then calculated as the area under each PR curve for every class, under each of these settings. We then take the mean of these APs, across the different classes, to get the mAP. The IoU threshold of 0.5, is used for the mAP 50 metric, while the range of IoU thresholds from 0.5 to 0.95 (in steps of 0.05) is used for the mAP 50-95 metric.  
%\KP{The meaning of mAP 50 and mAP 50-95 should be introduced here because you report such metrics in Table 3-5.}

\noindent \textbf{Implementation details:}
We train all models for $200$ epochs, with mini-batch size of $40$ images, where the gradients are accumulated over $2$ mini-batch iterations prior to parameter update. We use the ADAM optimizer~\cite{kingma2014adam} with a learning rate of $10^{-5}$ when training on the RGB modality and $10^{-3}$ when training on the IR modality. We use ``reduce on plateau" as the learning rate scheduler, that reduces learning rate by a factor of $0.1$ if the validation loss does not improve over $10$ epochs. Rather than initializing the RGB network from scratch, we initialize it with the pre-trained weights for detecting objects in the MS-COCO dataset~\cite{lin2014microsoft}. When explicitly encouraging complementarity between the RGB and IR branches, we set the weight $\omega_c= 0.01$ in Eq.~\ref{eq:final} such that both terms have comparable range.
%we weigh the objective in Eq.~\ref{eq:reg} with $0.01$.

%For pre-training we use MS COCO weights for baseline method and decomposed MS COCO pre-trained weights for proposed methods.

\subsection{Results and analysis}

\begin{table}[!t]
\centering
%\fontsize{6}{8.5}\selectfont
%\begin{tabularx}{\linewidth}{|X||r||r||r||r|}
\resizebox{\columnwidth}{!}{
\begin{tabular}{@{}c@{\hspace{2mm}}c@{\hspace{3mm}}c@{\hspace{3mm}}c@{\hspace{3mm}}c@{}}
\toprule
\textbf{Model} & \textbf{\# Parameters}$\downarrow$ & \textbf{Compression ($\%$)}$\uparrow$ & \textbf{mAP 50}$\uparrow$ & \textbf{mAP 50-95}$\uparrow$\\
\midrule
Baseline  & 37,205,480 &0  & $0.6826$ & \textbf{0.3173}\\
% Proposed ($\alpha=1$)  & $37,207,343$ & $-0.0050$ & $0.6898$ & $0.3182$\\
% \hline
\emph{\name} ($\alpha=0.9$)  & 35,400,800 & $4.8506$ & \textbf{0.6948} & $0.3162$\\
\emph{\name} ($\alpha=0.8$)  & \textbf{33,594,257} & \textbf{9.7062} & $0.6879$ & $0.3168$\\
\bottomrule
\end{tabular}
}
\caption{Results for FLIR Aligned RGB validation dataset.\label{tab:flir_aligned_results}}
\end{table}

\begin{table}[!t]
\centering
%\fontsize{6}{8.5}\selectfont
%\begin{tabularx}{\linewidth}{|X||r||r||r||r|}
\resizebox{\columnwidth}{!}{
\begin{tabular}{@{}c@{}c@{\hspace{3mm}}c@{\hspace{3mm}}c@{\hspace{3mm}}c@{}}
\toprule
\textbf{Model} & \textbf{\# Trainable Params}$\downarrow$ & \textbf{Compression ($\%$)}$\uparrow$ & \textbf{mAP 50}$\uparrow$ & \textbf{mAP 50-95}$\uparrow$\\
\midrule
Baseline  & 37,205,480 &0  & $0.5849$ & $0.2807$\\
% Proposed ($\alpha=1$)  & $37,207,343$ & $-0.0050$ & $0.5952$ & $0.2825$\\
% \hline
\emph{\name} ($\alpha=0.1$)  & \textbf{1,856,343} & \textbf{95.01} & $0.6205$ & \textbf{0.2807}\\
\emph{\name} ($\alpha=0.2$)  & 3,662,886 & $90.16$ & \textbf{0.6213} & $0.2794$\\
\bottomrule
\end{tabular}
}
\caption{Results for FLIR ADAS v1 IR validation dataset.\label{tab:flir_adas_v1_results}}
\end{table}

In Table \ref{tab:flir_aligned_results}, we present the evaluation results of the proposed and baseline methods for the FLIR Aligned RGB validation dataset for the task of object detection in RGB images. From the table, we observe that our proposed method demonstrates comparable, if not superior, performance to the baseline model in terms of both mAP 50 and mAP 50-95 evaluation metrics, across varying values of $\alpha$. Interestingly, while reduction in the value of $\alpha$ leads to significant compression in the model's size, our proposed method successfully maintains, and in certain instances enhances, model performance. We hypothesize that this is because the decrease of the trainable parameters reduces the chance of over-fitting. 
%This attests to the effectiveness of our proposed method.
%while substantially reducing the number of model parameters.
% performance metrics either improve or remain stable. This indicates that 
% However, a point for further scrutiny arises at $\alpha=1$, where a slight increase in the number of parameters is observed, leading to minuscule negative compression.

{Table \ref{tab:flir_adas_v1_results} presents the comparison results between the baseline and the proposed \emph{\name} method on the FLIR ADAS v1 IR validation dataset. For the proposed \emph{\name} method, we employ two different $\alpha$ configurations, $0.1$ and $0.2$, such that the ratios $\{\Delta r_l:r_l\}_{l=1}^L$ are $1:9$ and $1:4$, respectively, for $l=1,2,\ldots,L$. We observe that both proposed model configurations outperform the baseline on the mAP 50 evaluation metric, with only a few additional trainable parameters in the IR branch.
%, especially for smaller values of $\alpha$. 
These results underscore the potential of our proposed method to efficiently learn and generalize with significantly fewer trainable parameters in a data-scarce environment like the IR modality, while leveraging cross-modal cues from the data-rich RGB modality.}

Lastly, in Table \ref{tab:flir_adas_v1_abl_results}, we present results for augmenting the training objective with an explicit complementarity criterion, for $\alpha=0.1$ on the FLIR ADAS v1 IR validation dataset to determine the impact of regularization to promote learning of complementary features for the IR modality beyond the pre-trained RGB modality. 
%It is done by maximizing the distance between feature map distributions from RGB and IR branch using regularization terms. 
We observe that both $L_1$ and $L_2$ regularization methods show slight improvements in detection performance compared to the model without explicit regularization.

\begin{table}[!t]
\centering
\footnotesize
% \resizebox{\linewidth}{!}{%
%\begin{tabularx}{\linewidth}{|X||Y||Y|}
\begin{tabular}{ccc}
\toprule
\textbf{Regularization} & \textbf{mAP 50}$\uparrow$ & \textbf{mAP 50-95}$\uparrow$\\
\midrule
none & $0.6205$ & $0.2807$\\
$L_1$ & \textbf{0.6234} &  \textbf{0.2823}\\
$L_2$ & $0.6222$ & $0.2815$\\
\bottomrule
%KL Divergence & $0.6204$ & $0.2758$\\
%\hline
\end{tabular}
\caption{Results for \emph{\name} with explicit complementarity regularization for $\alpha=0.1$ on FLIR ADAS v1 IR validation dataset.\label{tab:flir_adas_v1_abl_results}}
\end{table}

% \noindent \textbf{Ablation studies:} 

\textbf{Qualitative results:}
In Figure \ref{results_1}, we compare the object detection results using the ground truth (left column), baseline (middle column), and proposed methods ($\alpha=0.1$, right column). The results are displayed vertically for two different images from the FLIR ADAS v1 IR validation dataset. We observe that the baseline method fails to detect small, distant objects and objects with backgrounds of similar texture as the foreground, whereas the proposed method accurately detects them. This shows that the proposed method is more robust against false negatives relative to the baseline.
Next, in Figure \ref{results_2}, we compare the object detection results for the proposed method without (left column) and with (right column) $L_1$ regularization and observe that this explicit regularization leads to more confident bounding box detections.

%% file: conclude.tex
\section{Conclusions}
\label{sec:conclude}

In this work, we proposed \emph{\name} -- a novel approach for object detection to be able to capture cross-modal cues so as to generalize better to modalities with scarce training data. \emph{\name} benefits from pre-training on modalities where plenty of training data is available (such as RGB), mitigating the challenges posed by the target modality (such as IR). In our formulation, at first, the data-rich RGB modality is used to learn the common cross-modal cues using low-rank tensor factorization of the network weights. We then use the IR modality training data to only learn the cues complementary to the RGB modality (either explicitly or implicitly), thereby requiring fewer trainable parameters.  We empirically validate the efficacy of our method on the task of object detection in IR images by pre-training our network on RGB object detection datasets and show that \emph{\name} yields performance boosts for object detection, in both RGB and IR images without an increase in the total number of network parameters.

%% file: domainadapt_iccvw23.bbl
\begin{thebibliography}{10}\itemsep=-1pt

\bibitem{ahmed2022cross}
Sk~Miraj Ahmed, Suhas Lohit, Kuan-Chuan Peng, Michael~J Jones, and Amit~K
  Roy-Chowdhury.
\newblock Cross-modal knowledge transfer without task-relevant source data.
\newblock In {\em European Conference on Computer Vision}, pages 111--127.
  Springer, 2022.

\bibitem{arthur2007k}
David Arthur and Sergei Vassilvitskii.
\newblock K-means++ the advantages of careful seeding.
\newblock In {\em Proceedings of the eighteenth annual ACM-SIAM symposium on
  Discrete algorithms}, pages 1027--1035, 2007.

\bibitem{bochkovskiy2020yolov4}
Alexey Bochkovskiy, Chien-Yao Wang, and Hong-Yuan~Mark Liao.
\newblock Y{OLO}v4: Optimal speed and accuracy of object detection.
\newblock {\em arXiv preprint arXiv:2004.10934}, 2020.

\bibitem{campbell2002biological}
Angela~L Campbell, Rajesh~R Naik, Laura Sowards, and Morley~O Stone.
\newblock Biological infrared imaging and sensing.
\newblock {\em Micron}, 33(2):211--225, 2002.

\bibitem{cao2019every}
Yu Cao, Tong Zhou, Xinhua Zhu, and Yan Su.
\newblock Every feature counts: An improved one-stage detector in thermal
  imagery.
\newblock In {\em 2019 IEEE 5th International Conference on Computer and
  Communications (ICCC)}, pages 1965--1969. IEEE, 2019.

\bibitem{dai2021tirnet}
Xuerui Dai, Xue Yuan, and Xueye Wei.
\newblock T{IRN}et: Object detection in thermal infrared images for autonomous
  driving.
\newblock {\em Applied Intelligence}, 51:1244--1261, 2021.

\bibitem{devaguptapu2019borrow}
Chaitanya Devaguptapu, Ninad Akolekar, Manuj M~Sharma, and Vineeth
  N~Balasubramanian.
\newblock Borrow from anywhere: Pseudo multi-modal object detection in thermal
  imagery.
\newblock In {\em Proceedings of the IEEE/CVF Conference on Computer Vision and
  Pattern Recognition Workshops}, pages 0--0, 2019.

\bibitem{dhanaraj2020vehicle}
Mayur Dhanaraj, Manish Sharma, Tiyasa Sarkar, Srivallabha Karnam, Dimitris~G
  Chachlakis, Raymond Ptucha, Panos~P Markopoulos, and Eli Saber.
\newblock Vehicle detection from multi-modal aerial imagery using {YOLO}v3 with
  mid-level fusion.
\newblock In {\em Big data II: learning, analytics, and applications}, volume
  11395, pages 22--32. SPIE, 2020.

\bibitem{donahue2013semi}
Jeff Donahue, Judy Hoffman, Erik Rodner, Kate Saenko, and Trevor Darrell.
\newblock Semi-supervised domain adaptation with instance constraints.
\newblock In {\em Proceedings of the IEEE conference on computer vision and
  pattern recognition}, pages 668--675, 2013.

\bibitem{du2021fa}
Shuangjiang Du, Baofu Zhang, Pin Zhang, Peng Xiang, and Hong Xue.
\newblock F{A-YOLO}: An improved {YOLO} model for infrared occlusion object
  detection under confusing background.
\newblock {\em Wireless Communications and Mobile Computing}, 2021:1--10, 2021.

\bibitem{dubey2018coreset}
Abhimanyu Dubey, Moitreya Chatterjee, and Narendra Ahuja.
\newblock Coreset-based neural network compression.
\newblock In {\em Proceedings of the European Conference on Computer Vision
  (ECCV)}, pages 454--470, 2018.

\bibitem{erkent2020semantic}
{\"O}zg{\"u}r Erkent and Christian Laugier.
\newblock Semantic segmentation with unsupervised domain adaptation under
  varying weather conditions for autonomous vehicles.
\newblock {\em IEEE Robotics and Automation Letters}, 5(2):3580--3587, 2020.

\bibitem{flir_aligned}
{FLIR aligned}.
\newblock {FLIR Aligned Dataset}, 2020.
\newblock {Accessed: August 20, 2022}.

\bibitem{ghose2019pedestrian}
Debasmita Ghose, Shasvat~M Desai, Sneha Bhattacharya, Deep Chakraborty,
  Madalina Fiterau, and Tauhidur Rahman.
\newblock Pedestrian detection in thermal images using saliency maps.
\newblock In {\em Proceedings of the IEEE/CVF Conference on Computer Vision and
  Pattern Recognition Workshops}, pages 0--0, 2019.

\bibitem{girshick2015fast}
Ross Girshick.
\newblock Fast {R-CNN}.
\newblock In {\em Proceedings of the IEEE international conference on computer
  vision}, pages 1440--1448, 2015.

\bibitem{girshick2014rich}
Ross Girshick, Jeff Donahue, Trevor Darrell, and Jitendra Malik.
\newblock Rich feature hierarchies for accurate object detection and semantic
  segmentation.
\newblock In {\em Proceedings of the IEEE conference on computer vision and
  pattern recognition}, pages 580--587, 2014.

\bibitem{guan2021uncertainty}
Dayan Guan, Jiaxing Huang, Aoran Xiao, Shijian Lu, and Yanpeng Cao.
\newblock Uncertainty-aware unsupervised domain adaptation in object detection.
\newblock {\em IEEE Transactions on Multimedia}, 24:2502--2514, 2021.

\bibitem{he2017mask}
Kaiming He, Georgia Gkioxari, Piotr Doll{\'a}r, and Ross Girshick.
\newblock Mask {R-CNN}.
\newblock In {\em Proceedings of the IEEE international conference on computer
  vision}, pages 2961--2969, 2017.

\bibitem{he2016deep}
Kaiming He, Xiangyu Zhang, Shaoqing Ren, and Jian Sun.
\newblock Deep residual learning for image recognition.
\newblock In {\em Proceedings of the IEEE conference on computer vision and
  pattern recognition}, pages 770--778, 2016.

\bibitem{herrmann2018cnn}
Christian Herrmann, Miriam Ruf, and J{\"u}rgen Beyerer.
\newblock C{NN}-based thermal infrared person detection by domain adaptation.
\newblock In {\em Autonomous Systems: Sensors, Vehicles, Security, and the
  Internet of Everything}, volume 10643, pages 38--43. SPIE, 2018.

\bibitem{hsu2020progressive}
Han-Kai Hsu, Chun-Han Yao, Yi-Hsuan Tsai, Wei-Chih Hung, Hung-Yu Tseng, Maneesh
  Singh, and Ming-Hsuan Yang.
\newblock Progressive domain adaptation for object detection.
\newblock In {\em Proceedings of the IEEE/CVF winter conference on applications
  of computer vision}, pages 749--757, 2020.

\bibitem{hu2021lora}
Edward~J Hu, Phillip Wallis, Zeyuan Allen-Zhu, Yuanzhi Li, Shean Wang, Lu Wang,
  Weizhu Chen, et~al.
\newblock L{oRA}: Low-rank adaptation of large language models.
\newblock In {\em International Conference on Learning Representations}, 2021.

\bibitem{kamalakara2022exploring}
Siddhartha~Rao Kamalakara, Acyr Locatelli, Bharat Venkitesh, Jimmy Ba, Yarin
  Gal, and Aidan~N Gomez.
\newblock Exploring low rank training of deep neural networks.
\newblock {\em arXiv preprint arXiv:2209.13569}, 2022.

\bibitem{kieu2020task}
My Kieu, Andrew~D Bagdanov, Marco Bertini, and Alberto Del~Bimbo.
\newblock Task-conditioned domain adaptation for pedestrian detection in
  thermal imagery.
\newblock In {\em European Conference on Computer Vision}, pages 546--562.
  Springer, 2020.

\bibitem{kingma2014adam}
Diederik~P Kingma and Jimmy Ba.
\newblock Adam: A method for stochastic optimization.
\newblock {\em arXiv preprint arXiv:1412.6980}, 2014.

\bibitem{krivsto2020thermal}
Mate Kri{\v{s}}to, Marina Ivasic-Kos, and Miran Pobar.
\newblock Thermal object detection in difficult weather conditions using
  {YOLO}.
\newblock {\em IEEE access}, 8:125459--125476, 2020.

\bibitem{li2020research}
M~e Li, Tao Zhang, and W Cui.
\newblock Research of infrared small pedestrian target detection based on
  {YOLO}v3.
\newblock {\em Infrared Technol}, 42:176--181, 2020.

\bibitem{li2021category}
Shuai Li, Jianqiang Huang, Xian-Sheng Hua, and Lei Zhang.
\newblock Category dictionary guided unsupervised domain adaptation for object
  detection.
\newblock In {\em Proceedings of the AAAI conference on artificial
  intelligence}, volume~35, pages 1949--1957, 2021.

\bibitem{li2021yolo}
Shasha Li, Yongjun Li, Yao Li, Mengjun Li, and Xiaorong Xu.
\newblock Y{OLO-FIRI}: Improved {YOLO}v5 for infrared image object detection.
\newblock {\em IEEE access}, 9:141861--141875, 2021.

\bibitem{li2021infrared}
Wei Li.
\newblock Infrared image pedestrian detection via yolo-v3.
\newblock In {\em 2021 IEEE 5th Advanced Information Technology, Electronic and
  Automation Control Conference (IAEAC)}, volume~5, pages 1052--1055. IEEE,
  2021.

\bibitem{li2020yolo}
Yongjun Li, Shasha Li, Haohao Du, Lijia Chen, Dongming Zhang, and Yao Li.
\newblock Y{OLO-ACN}: Focusing on small target and occluded object detection.
\newblock {\em IEEE access}, 8:227288--227303, 2020.

\bibitem{li2017scene}
Yikang Li, Wanli Ouyang, Bolei Zhou, Kun Wang, and Xiaogang Wang.
\newblock Scene graph generation from objects, phrases and region captions.
\newblock In {\em Proceedings of the IEEE international conference on computer
  vision}, pages 1261--1270, 2017.

\bibitem{lin2014microsoft}
Tsung-Yi Lin, Michael Maire, Serge Belongie, James Hays, Pietro Perona, Deva
  Ramanan, Piotr Doll{\'a}r, and C~Lawrence Zitnick.
\newblock Microsoft {COCO}: Common objects in context.
\newblock In {\em Computer Vision--ECCV 2014: 13th European Conference, Zurich,
  Switzerland, September 6-12, 2014, Proceedings, Part V 13}, pages 740--755.
  Springer, 2014.

\bibitem{manssor2022real}
Samah~AF Manssor, Shaoyuan Sun, Mohammed Abdalmajed, and Shima Ali.
\newblock Real-time human detection in thermal infrared imaging at night using
  enhanced {T}iny-yolov3 network.
\newblock {\em Journal of Real-Time Image Processing}, pages 1--14, 2022.

\bibitem{ouyang2013joint}
Wanli Ouyang and Xiaogang Wang.
\newblock Joint deep learning for pedestrian detection.
\newblock In {\em Proceedings of the IEEE international conference on computer
  vision}, pages 2056--2063, 2013.

\bibitem{redmon2016you}
Joseph Redmon, Santosh Divvala, Ross Girshick, and Ali Farhadi.
\newblock You only look once: Unified, real-time object detection.
\newblock In {\em Proceedings of the IEEE conference on computer vision and
  pattern recognition}, pages 779--788, 2016.

\bibitem{redmon2018yolov3}
Joseph Redmon and Ali Farhadi.
\newblock Y{OLO}v3: An incremental improvement.
\newblock {\em arXiv preprint arXiv:1804.02767}, 2018.

\bibitem{ren2015faster}
Shaoqing Ren, Kaiming He, Ross Girshick, and Jian Sun.
\newblock Faster {R-CNN}: Towards real-time object detection with region
  proposal networks.
\newblock {\em Advances in neural information processing systems}, 28, 2015.

\bibitem{rodriguez2019domain}
Adrian~Lopez Rodriguez and Krystian Mikolajczyk.
\newblock Domain adaptation for object detection via style consistency.
\newblock {\em arXiv preprint arXiv:1911.10033}, 2019.

\bibitem{sharma2020yolors}
Manish Sharma, Mayur Dhanaraj, Srivallabha Karnam, Dimitris~G Chachlakis,
  Raymond Ptucha, Panos~P Markopoulos, and Eli Saber.
\newblock Y{OLO}rs: Object detection in multimodal remote sensing imagery.
\newblock {\em IEEE Journal of Selected Topics in Applied Earth Observations
  and Remote Sensing}, 14:1497--1508, 2020.

\bibitem{sharma2021yolors}
Manish Sharma, Panos~P Markopoulos, and Eli Saber.
\newblock Yolors-lite: A lightweight cnn for real-time object detection in
  remote-sensing.
\newblock In {\em 2021 IEEE International Geoscience and Remote Sensing
  Symposium IGARSS}, pages 2604--2607. IEEE, 2021.

\bibitem{sharma2021convolutional}
Manish Sharma, Panos~P Markopoulos, Eli Saber, M~Salman Asif, and Ashley
  Prater-Bennette.
\newblock Convolutional auto-encoder with tensor-train factorization.
\newblock In {\em Proceedings of the IEEE/CVF international conference on
  computer vision}, pages 198--206, 2021.

\bibitem{shuigen2021infrared}
Wei Shuigen, Wang Chengwei, Chen Zhen, Z Congxuan, and Z Xiaoyu.
\newblock Infrared dim target detection based on human visual mechanism.
\newblock {\em Acta Photonica Sinica}, 50(1):0110001, 2021.

\bibitem{singh2023multimodal}
Saurav Singh, Manish Sharma, Jamison Heard, Jesse~D Lew, Eli Saber, and Panos~P
  Markopoulos.
\newblock Multimodal aerial view object classification with disjoint unimodal
  feature extraction and fully-connected-layer fusion.
\newblock In {\em Big Data V: Learning, Analytics, and Applications}, volume
  12522, page 1252206. SPIE, 2023.

\bibitem{song2021multispectral}
Xiaoru Song, Song Gao, and Chaobo Chen.
\newblock A multispectral feature fusion network for robust pedestrian
  detection.
\newblock {\em Alexandria Engineering Journal}, 60(1):73--85, 2021.

\bibitem{sun2017revisiting}
Chen Sun, Abhinav Shrivastava, Saurabh Singh, and Abhinav Gupta.
\newblock Revisiting unreasonable effectiveness of data in deep learning era.
\newblock In {\em Proceedings of the IEEE international conference on computer
  vision}, pages 843--852, 2017.

\bibitem{sun2021sparse}
Peize Sun, Rufeng Zhang, Yi Jiang, Tao Kong, Chenfeng Xu, Wei Zhan, Masayoshi
  Tomizuka, Lei Li, Zehuan Yuan, Changhu Wang, et~al.
\newblock Sparse {R-CNN}: End-to-end object detection with learnable proposals.
\newblock In {\em Proceedings of the IEEE/CVF conference on computer vision and
  pattern recognition}, pages 14454--14463, 2021.

\bibitem{sun2015deepid3}
Yi Sun, Ding Liang, Xiaogang Wang, and Xiaoou Tang.
\newblock Deep{ID}3: Face recognition with very deep neural networks.
\newblock {\em arXiv preprint arXiv:1502.00873}, 2015.

\bibitem{flir_adas_v1}
{Teledyne Technologies Incorporated}.
\newblock {FLIR ADAS v1 Dataset}, 2020.
\newblock {Accessed: August 20, 2022}.

\bibitem{ustinova2016learning}
Evgeniya Ustinova and Victor Lempitsky.
\newblock Learning deep embeddings with histogram loss.
\newblock {\em Advances in neural information processing systems}, 29, 2016.

\bibitem{wang2023yolov7}
Chien-Yao Wang, Alexey Bochkovskiy, and Hong-Yuan~Mark Liao.
\newblock {Y{OLO}v7}: Trainable bag-of-freebies sets new state-of-the-art for
  real-time object detectors.
\newblock In {\em Proceedings of the IEEE/CVF Conference on Computer Vision and
  Pattern Recognition}, pages 7464--7475, 2023.

\bibitem{wang2023ssda3d}
Yan Wang, Junbo Yin, Wei Li, Pascal Frossard, Ruigang Yang, and Jianbing Shen.
\newblock S{SDA3D}: Semi-supervised domain adaptation for 3{D} object detection
  from point cloud.
\newblock In {\em Proceedings of the AAAI Conference on Artificial
  Intelligence}, volume~37, pages 2707--2715, 2023.

\bibitem{wei2020incremental}
Xing Wei, Shaofan Liu, Yaoci Xiang, Zhangling Duan, Chong Zhao, and Yang Lu.
\newblock Incremental learning based multi-domain adaptation for object
  detection.
\newblock {\em Knowledge-Based Systems}, 210:106420, 2020.

\bibitem{yao2021multi}
Xingxu Yao, Sicheng Zhao, Pengfei Xu, and Jufeng Yang.
\newblock Multi-source domain adaptation for object detection.
\newblock In {\em Proceedings of the IEEE/CVF International Conference on
  Computer Vision}, pages 3273--3282, 2021.

\bibitem{yu2019unsupervised}
Fuxun Yu, Di Wang, Yinpeng Chen, Nikolaos Karianakis, Tong Shen, Pei Yu,
  Dimitrios Lymberopoulos, Sidi Lu, Weisong Shi, and Xiang Chen.
\newblock Unsupervised domain adaptation for object detection via cross-domain
  semi-supervised learning.
\newblock {\em arXiv preprint arXiv:1911.07158}, 2019.

\bibitem{zhang2022multi}
Dan Zhang, Mao Ye, Yiguang Liu, Lin Xiong, and Lihua Zhou.
\newblock Multi-source unsupervised domain adaptation for object detection.
\newblock {\em Information Fusion}, 78:138--148, 2022.

\bibitem{zhang2017fcn}
Shanghang Zhang, Guanhang Wu, Joao~P Costeira, and Jos{\'e}~MF Moura.
\newblock F{CN}-r{LSTM}: Deep spatio-temporal neural networks for vehicle
  counting in city cameras.
\newblock In {\em Proceedings of the IEEE international conference on computer
  vision}, pages 3667--3676, 2017.

\bibitem{zhao2021iyolo}
Xiaofeng Zhao, Yebin Xu, Fei Wu, Wei Cai, and Zhili Zhang.
\newblock I{YOLO}: Multi-scale infrared target detection method based on
  bidirectional feature fusion.
\newblock In {\em Journal of Physics: Conference Series}, volume 1873, page
  012020. IOP Publishing, 2021.

\bibitem{zhou2022multi}
Wenzhang Zhou, Dawei Du, Libo Zhang, Tiejian Luo, and Yanjun Wu.
\newblock Multi-granularity alignment domain adaptation for object detection.
\newblock In {\em Proceedings of the IEEE/CVF Conference on Computer Vision and
  Pattern Recognition}, pages 9581--9590, 2022.

\end{thebibliography}
